\documentclass[conference]{IEEEtran}
\IEEEoverridecommandlockouts
\usepackage{cite}
\usepackage{amsmath,amssymb,amsfonts}
\usepackage[noend]{algpseudocode}
\usepackage{graphicx}
\usepackage{textcomp}
\usepackage{xcolor}
\usepackage{hyperref}
\def\BibTeX{{\rm B\kern-.05em{\sc i\kern-.025em b}\kern-.08em
    T\kern-.1667em\lower.7ex\hbox{E}\kern-.125emX}}
\begin{document}

\title{Cross-Domain  Ambiguity  Detection  using  Linear Transformation of Word Embedding Spaces
}

\author{\IEEEauthorblockN{Vaibhav Jain\IEEEauthorrefmark{1}, Ruchika Malhotra, Sanskar Jain, and Nishant Tanwar}\IEEEauthorblockA{Department of Software Engineering\\Delhi Technological University, Delhi, India}\IEEEauthorblockA{\IEEEauthorrefmark{1}Email: vaibhav29498@gmail.com}}

\maketitle

\begin{abstract}
The requirements engineering process is a crucial stage of the software development life cycle. It involves various stakeholders from different professional backgrounds, particularly in the requirements elicitation phase. Each stakeholder carries distinct domain knowledge, causing them to differently interpret certain words, leading to cross-domain ambiguity. This can result in misunderstanding amongst them and jeopardize the entire project. This paper proposes a natural language processing approach to find potentially ambiguous words for a given set of domains. The idea is to apply linear transformations on word embedding models trained on different domain corpora, to bring them into a \textit{unified embedding space}. The approach then finds words with divergent embeddings as they signify a variation in the meaning across the domains. It can help a requirements analyst in preventing misunderstandings during elicitation interviews and meetings by defining a set of potentially ambiguous terms in advance. The paper also discusses certain problems with the existing approaches and discusses how the proposed approach resolves them.
\end{abstract}

\begin{IEEEkeywords}
cross-domain ambiguity, linear transformation, requirements engineering, word embeddings, natural language processing
\end{IEEEkeywords}

\section{Introduction}
In the context of software engineering, requirements engineering is the process of describing the intended behaviour of a software system along with the associated constraints~\cite{pressman2010software}. One of its phase is requirements elicitation, which has been termed as the most difficult, critical, and communication-intensive aspect of software development~\cite{aggarwal2005software}. It requires interaction between different stakeholders through various techniques like brainstorming sessions and facilitated application specification technique. A stakeholder is any person with a vested interest in the project, such as potential users, developers, testers, domain experts, and regulatory agency personnel~\cite{singh2012object}. As these stakeholders come from different professional backgrounds and carry different domain knowledge, cross-domain ambiguity can occur amongst them. One may assign an interpretation to another's expression different from the intended meaning. This results in misunderstanding and distrust in requirements elicitation meetings, and costly problems in the later stages of the software life cycle~\cite{wang2013}.

The study of variation of word meanings across domains as an NLP problem is termed as \textit{Synchronic Lexical Semantic Change (LSC) Detection}~\cite{sch2019}. The first attempt to apply it for dealing with cross-domain ambiguity in requirements engineering was by Ferrari \textit{et al.} (2017) who used Wikipedia crawling and word embeddings to estimate ambiguous computer science (CS) terms vis-à-vis other application domains~\cite{ferrari2017}. Mishra and Sharma extended this work by focusing on various engineering subdomains~\cite{mishra2019}. Another approach was suggested by Ferrari \textit{et al.} (2018) which also considered the ambiguity caused by non-CS domain-specific words and addressed some of the technical limitations of the previous work~\cite{ferrari2018}. This approach was later extended to include quantitative evaluation of the obtained results~\cite{ferrari2019}. An alternative approach which doesn't require domain-specific word embeddings was suggested by Toews and Holland~\cite{toews}.

This paper proposes a natural language processing (NLP) approach based on linear transformation of word embedding spaces. Word embedding is a vector representation of a word capable of capturing its semantic and syntactic relations. A linear transformation can be used to learn a linear relationship between two word embedding spaces. The proposed approach produces a ranked list of potentially ambiguous terms for a given set of domains. It constructs a word embedding space for each domain using corpora composed of Wikipedia articles. It then applies linear transformations on these spaces in order to align them and construct a \textit{unified embedding space}. For each word in a set of target words, an ambiguity score is assigned by applying a distance metric on its \textit{domain-specific embeddings}.

The remainder of this paper is organised as follows: Section \ref{sec:prelims} provides some background on word embeddings and linear transformation of embedding spaces. The existing approaches to cross-domain ambiguity detection are briefly explained in Section \ref{sec:related}. The motivation behind the proposed approach is discussed in Section \ref{sec:motivation}, whereas the apporach itself is outlined in Section \ref{sec:approach}. The experimental setup and results are presented and discussed in Section \ref{results}, and the conclusion and details of planned future work are provided in Section \ref{sec:final}.

\section{Preliminaries}
\label{sec:prelims}
\subsection{Word Embeddings}
Word embedding is a collective term for language modelling techniques that map each word in the vocabulary to a dense vector representation. Contrary to one-hot representation, word embedding techniques embed each word into a low-dimensional continuous space and capture its semantic and syntactic relationships~\cite{li2015}. It is based on the distributional hypothesis proposed by Harris which states that words appearing in similar linguistic contexts share similar meanings~\cite{harris1954}.

One of the most popular word embedding techniques is skip-gram with negative sampling (SGNS)~\cite{mikolov2013}. It trains a shallow two-layer neural network which, given a single input word $w$, predicts a set of context words $c(w)$. The context for a word $w_i$ is the set of words surrounding it in a fixed-size window, i.e. $\{w_{i-L},\cdots,w_{i-1},w_{i+1},\cdots,w_{i+L}\}$, where $L$ is the context-window size. Each word $w$ is associated with vectors $u_w \in \mathbb{R}^D$ and $v_w \in \mathbb{R}^D$, called the input and output vectors respectively. If $T$ is the number of windows in the given corpus, then the objective of the skip-gram model is to maximize 
\begin{equation}
    \frac{1}{T}\sum_{t=1}^T\sum_{-L \leqslant i\leqslant L; i \neq 0}\log{p(w_{t+i}|w_t)}
\end{equation}
In the negative sampling method, $p(w_{t+i}|w_i)$ is defined as
\begin{equation}
    p(w_O|w_I) = \log{\sigma(u_{w_I}^Tv_{w_O})} + \sum_{i=1}^k \log{\sigma(-u_{w_I}^Tv_{w_i})}
\end{equation}
where $w_i\sim P(w)$ and $P(w)$ is the noise distribution.

\subsection{Linear Transformation}

A linear transformation can be used to learn a linear mapping from one vector space to another. Its use for combining different word embedding spaces was first explored by Mikolov \textit{et al.} who used it for bilingual machine translation~\cite{mikolov2013exploiting}. They used a list of word pairs $\{x_i,y_i\}_{i=1}^n$, where $y_i$ is the translation of $x_i$. Then they learned a \textit{translation matrix} $W$ by minimizing the following loss function
\begin{equation}
    \sum_{i=1}^n |x_iW - y_i|
\end{equation}
This approach can also be used for aligning monolingual word embeddings. If one assumes that the meaning of most words remains unchanged, linear regression can be used to find the best rotational alignment between two word embedding spaces. Failure to properly align a word can be then used to identify a change in meaning. This is the basis for the proposed approach towards identifying cross-domain ambiguous words. Similar approaches have been used to detect linguistic variation in the meaning of a word with time and to develop ensemble word embedding models~\cite{hamilton,kulkarni,muromagi}.

Significant work has been done to improve the linear transformation method. Dimension-wise mean centering has been shown to improve the performance of linear transformation methods in downstream tasks~\cite{artetxe2016}. Xing \textit{et al.} noticed a hypothetical inconsistency in the distance metrics used in the optimization objectives in the work of Mikolov \textit{et al.}: dot product for training word embeddings, Euclidean distance for learning transformation matrix, and cosine distance for similarity computations~\cite{xing2015}. It was solved by normalizing the word embeddings and by requiring the transformation matrix to be orthogonal. The optimal orthogonal transformation matrix which maps $X$ to $Y$ can be found through the solution of the well-known Orthogonal Procrustes problem, which is given by
\begin{equation}
    W = UV^T
    \label{eq:svd}
\end{equation}
where $X^TY = U\Sigma V^T$ is the singular value decomposition (SVD) factorization of $X^TY$~\cite{Sch1966}.

\section{Related Work}
\label{sec:related}
Synchronic LSC detection refers to the measurement of variation of word meanings across domains or speaker communities~\cite{sch2019}. The latter has been studied by making use of the large-scale data provided by communities on online platforms such as Reddit~\cite{tredici-fernandez-2017-semantic}.

Research works on cross-domain ambiguity detection have been limited to its applicability in requirements engineering. The first approach was suggested by Ferrari \textit{et al.} (2017) who employed Wikipedia crawling and word embeddings to estimate the variation of typical CS words (e.g., code, database, windows) in other domains~\cite{ferrari2017}. They used Wikipedia articles to create two corpora: a CS one and a domain-specific one, replaced the target words (top-k most frequent nouns in the CS corpus) in the latter by a uniquely identifiable modified version, and trained a single language model for both corpora. Cosine similarity was then used as a metric to estimate the variation in the meaning of the target words when they are used in the specified domain. However, this approach suffers from the following drawbacks: \begin{itemize}\item the inability to identify non-CS cross-domain ambiguous words, \item the need to construct a language model for each combination of domains, and \item the need to modify the domain-specific corpus.\end{itemize} Their work was extended by Mishra and Sharma who applied it on various subdomains of engineering with varying corpus size~\cite{mishra2019}. They identified the most suitable hyperparameters for training word embeddings on corpora of three different classes: large, medium, and small, based on the number of documents. They then used the obtained results to identify a similarity threshold for ambiguous words.

Ferrari \textit{et al.} (2018) suggested an approach based on developing word embedding spaces for each domain, and then estimating the variation in the meaning of a word by comparing the lists of its most similar words in each domain~\cite{ferrari2018}. This approach addressed the above-mentioned drawbacks of the previous one. It was later extended by Ferrari and Esuli, with the major contribution being the introduction of a quantitative evaluation of the approach~\cite{ferrari2019}.

An alternative approach which does not require domain-specific word embeddings was suggested by Toews and Holland~\cite{toews}. It estimates a word's similarity across domains through context similarity. This approach does require trained word embeddings, but they are not required to be domain-specific, which allows it to be used on small domain corpora as well. If $D_1$ and $D_2$ are two domain corpora, then the context similarity of a word $w$ is defined as
\begin{equation}
    simc(w) = \frac{center(c_1)\cdot center(c_2)}{\|center(c_1)\|\cdot\|center(c_2)\|}
\end{equation}
\begin{equation}
    center(c) = \frac{1}{|c|}\sum_{w\in c}IDF_D(w)\cdot v_w
\end{equation}
where $c_1\subset D_1$ and $c_2\subset D_2$ consist of all words from sentences containing $w$.

\section{Motivation}
\label{sec:motivation}
The motivation behind the proposed linear transformation based approach is based on the following factors:

\begin{itemize}
    \item The approaches suggested by Ferrari \textit{et al.} (2018) and Toews and Holland judge a word's meaning from its local context rather than a global one. This leads them to wrongly assign a high ambiguity score to a word having distinct, yet similar, nearest words in different domains. A particular example of this problem is the high score assigned to proper names such as \textit{Michael}; although they are near to other proper nouns in all domains, but the exact lists vary widely. Such approaches also fail in the opposite scenario in which the meaning of the nearest words themselves change. This can happen in the case of \textit{ambiguous clusters}. For example, a lot of topics in artificial intelligence, such as neural networks and genetic algorithms, are inspired by biology. Due to this, certain words appear together in both these domains but carry different interpretations.
    However, the approach proposed by this paper relies on the global context rather than the local one, which resolves the issues mentioned above.
    \item The proposed approach can work for more than two domains as opposed to the approaches suggested by Ferrari \textit{et al.} (2017) and Toews and Holland~\cite{ferrari2017, toews}.
    \item The approach proposed by Ferrari \textit{et al.} (2018) assumes the meaning of the neighbouring words to be the same across domains, whereas a linear transformation based approach works on a much weaker assumption that the meaning of most words remains the same across domains.
    \item Schlechtweg \textit{et al.} evaluated various synchronic LSC detection models on \textit{SURel}, a German dataset consisting of the meaning variations from general to domain-specific corpus determined through manual annotation~\cite{hatty-etal-2019-surel,sch2019}. Their study found linear transformation to perform much better than other alignment techniques such as word injection (proposed by Ferrari \textit{et al.} (2017)) and vector initialization.
\end{itemize}

\section{Approach}
\label{sec:approach}
Given a set of domains $D = \{D_1, \cdots, D_n\}$, the approach requires a word embedding space $S_i$ corresponding to each domain $D_i$. The first step is to align the embedding spaces (subsection \ref{subsec:align}) and then determine the set of target words (subsection \ref{subsec:target}). The final step is to assign a cross-domain ambiguity score to each target word (subsection \ref{subsec:ranking}).

\subsection{Embedding Spaces Alignment}
\label{subsec:align}
This step determines a transformation matrix $M_i$ for each domain-specific word embedding space $S_i$ which maps it to a unified embedding space. It uses an algorithm devised by Murom{\"a}gi \textit{et al.}~\cite{muromagi} which iteratively finds the transformation matrices $M_1, M_2, \cdots, M_n$ and the common target space $Y$. It performs the following two steps in each iteration:

\begin{enumerate}
    \item The transformation matrices $M_1, M_2, \cdots, M_n$ are calculated using equation~\ref{eq:svd}.
    \item The target space is updated to be the average of all transformed spaces:
    \begin{equation}
        Y(w) = \frac{1}{n_w} \sum_{i=1}^{n_w} S_i(w)M_i
    \end{equation}
    where $n_w$ is the number of domain-specific embedding spaces with word $w$ as part of its vocabulary.
\end{enumerate}

These steps are repeatedly performed as long as the change in average normalised residual error, which is given by

\begin{equation}
    \frac{1}{n} \sum_{i=1}^n \frac{\| S_iM_i - Y\|}{\sqrt{|S_i|\cdot d}}
\end{equation}
is equal to or greater than a predefined threshold $\tau$.

\subsection{Target Words Selection}
\label{subsec:target}
The approach for identifying the set of target words $T_D$ has been presented in Figure~\ref{algo:terms}.

\begin{figure}[h]
\begin{algorithmic}[0]
    \Procedure{SelectWords}{$C, k, \rho$}
        \State $T_D\gets \emptyset$
        \For{$w_i\in \Call{Vocab}{C_1}\cup\cdots\cup\Call{Vocab}{C_n}$}
            \If{$\Call{POS}{w_i} \in \{NN, VB, ADJ\}$}
                \State $counts = \{\Call{Freq}{C_1, w_i}, \cdots, \Call{Freq}{C_n,w_i}$\}
                \State $c_1, c_2\gets\Call{Top2Values}{counts}$
                \If{$c_1\geqslant k\land c_2\geqslant \rho\times c_1$}
                    \State $T_D\gets T_D\cup\{w_i\}$
                \EndIf
            \EndIf
        \EndFor
        \State \textbf{return} $T_D$
    \EndProcedure
\end{algorithmic}
\caption{Algorithm for selecting target words}\label{algo:terms}
\end{figure}

This step requires two numerical parameters, $k$ and $\rho$. To be considered a target word, $w$ must satisfy three conditions:
\begin{enumerate}
    \item It must be a content word, i.e. noun, verb, or adjective.
    \item Its maximum frequency in a domain corpus, i.e. $f_{max} = max(count_i(w))$, should be greater than or equal to $k$.
    \item It should have a frequency of at least $\rho f_{max}$ in any other domain corpus. 
\end{enumerate}

\subsection{Cross-Domain Ambiguity Ranking}
\label{subsec:ranking}
This step assigns an \textit{ambiguity score} to each word in $T_D$ based on their cross-domain ambiguity across the corpora $C = \{C_1, \cdots, C_n\}$. The algorithm for the same is reported in Figure~\ref{algo:ranking}.

\begin{figure}[h]
\begin{algorithmic}[0]
    \Procedure{AssignAmbiguityScores}{$T_D, M, S$}
        \State $Score\gets \emptyset$
        \For{$w\in T_D$}
            \State $V\gets \emptyset$
            \For{$S_i\in S$}
                \If{$w\in S_i$}
                    \State $V\gets V\cup \{M_iS_i(w)\}$
                \EndIf
            \EndFor
            \State $U\gets 0$
            \State $C\gets 0$
            \For{$v_i\in V$}
                \For{$v_j\in V\setminus v_i$}
                    \State $c\gets count_i(w) + count_j(w)$
                    \State $U\gets U + c\times\Call{CosineDistance}{v_i, v_j}$
                    \State $C \gets C + c$
                \EndFor
            \EndFor
            \State $Score[w]\gets U / C$
        \EndFor
        \State $A_D\gets \Call{Sort}{T_D, Score}$
        \State \textbf{return} $A_D$
    \EndProcedure
\end{algorithmic}
\caption{Algorithm for assigning ambiguity scores}\label{algo:ranking}
\end{figure}

The idea is as follows. For each word $w$ in the set of target words $T_D$, the cosine distance for each unordered pair of its transformed embeddings is calculated, which is given by

\begin{equation}
    cosineDistance(v_i, v_j) = 1 - \frac{v_i\cdot v_j}{\|v_i\|\|v_j\|}
\end{equation}

The average of all these cosine distances, weighted by the sum of the word frequencies in the corresponding domain corpora, is the ambiguity score assigned to the word $w$. All words in $T_D$ are sorted according to their score and a ranked list $A_D$ is produced.

\section{Results}
\label{results}
\subsection{Project Scenarios}
To showcase the working of the proposed approach, this paper considers the same hypothetical project scenarios that were used by Ferrari and Esuli~\cite{ferrari2019}. They involve five domains: computer science (CS), electronic engineering (EE), mechanical engineering (ME), medicine (MED), and sports (SPO).

\begin{enumerate}
    \item \textit{Light Controller} [CS, EE]: an embedded software for room illumination system
    \item \textit{Mechanical CAD} [CS, ME]: a software for designing and drafting mechanical components.
    \item \textit{Medical Software} [CS, MED]: a disease-prediction software.
    \item \textit{Athletes Network} [CS, SPO]: a social network for athletes.
    \item \textit{Medical  Device} [CS, EE, MED]: a fitness tracker connected to a mobile app
    \item \textit{Medical  Robot} [CS, EE, ME, MED]: a computer-controlled robotic arm used for surgery.
    \item \textit{Sports Rehab Machine} [CS, EE, ME, MED, SPO]: a rehabilitation machine targeted towards athletes.
\end{enumerate}

The first four scenarios can be thought of as an interview between a requirements analyst with a CS and a domain expert, whereas the other three scenarios can be regarded as group elicitation meetings involving stakeholders from multiple domains.

\subsection{Experimental Setup}

\begin{figure}[t]
\centering
\includegraphics[scale=.35]{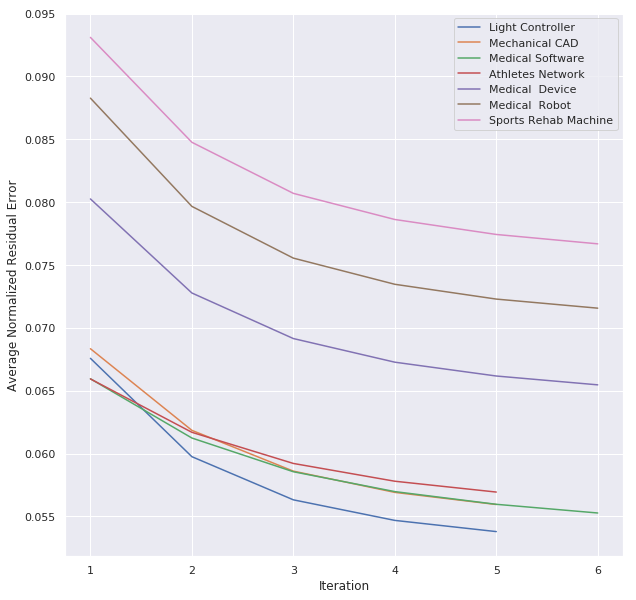}
\caption{Plot of the average normalized residual errors}
\label{fig:plot}
\end{figure}

The Wikipedia API for Python\footnote{\url{https://pypi.org/project/wikipedia/}} was used to construct the domain corpora by scraping articles belonging to particular categories. A maximum subcategory depth of 3 and a maximum article limit of 20,000 was set while creating each domain corpus.\footnote{Since \href{https://en.wikipedia.org/wiki/Category:Computer_science}{Category:Computer science} is a subcategory of \href{https://en.wikipedia.org/wiki/Category:Electronic_engineering}{Category:Electronic engineering}, it was excluded while creating the EE corpus to avoid extensive overlap with the CS corpus.} Each article text was converted to lowercase and all non-alphanumeric words and stop words were removed, followed by lemmatization. The article count, word count, and vocabulary size for each domain corpus are reported by Table~\ref{table:corpora}.

\begin{table}[htbp]
\caption{Domain Corpora Statistics}\label{table:corpora}
\begin{center}
\begin{tabular}{|l|c|c|c|}
\hline
Domain & Articles & Words & Vocabulary \\
\hline
Computer science & 20,000 & 80,37,521 & 1,77,764 \\
\hline
Electronic engineering & 16,420 & 77,10,843 & 1,79,898 \\
\hline
Mechanical engineering & 20,000 & 1,02,02,205 & 1,99,696 \\
\hline
Medicine & 20,000 & 80,45,379 & 2,00,266 \\
\hline
Sports & 20,000 & 94,48,453 &  2,42,583 \\
\hline
\end{tabular}
\end{center}
\end{table}

The word embeddings were trained using the \texttt{gensim}\footnote{\url{https://radimrehurek.com/gensim/}} implementation of the \texttt{word2vec} SGNS algorithm with word embedding dimension $d = 50$, context window size $L = 10$, negative sampling size $\eta = 5$, and minimum frequency $f_{min} = 10$. Training of the word embeddings was followed by length normalization and dimension-wise mean centering. For aligning the word embedding spaces, the threshold $\tau$ was set to $0.001$. The plot of average normalized residual errors for each project scenario is depicted in Figure~\ref{fig:plot}. The parameters for identifying target words were set as $k = 1000$ and $\rho = 0.5$.

\subsection{Cross-Domain Ambiguity Rankings}
The top-10 and bottom-10 ranked terms for each project scenario are reported along with their ambiguity scores by Table~\ref{table:rank}.

\begin{table*}
\caption{Ranked List of Target Words based on their Ambiguity Scores}\label{table:rank}
\centering
\begin{tabular}{|lc|lc|lc|lc|}
\hline
\multicolumn{2}{|c|}{Light controller} & \multicolumn{2}{c|}{Mechanical CAD} & \multicolumn{2}{c|}{Medical Software} & \multicolumn{2}{c|}{Athletes Network} \\
\hline
Term & Score & Term & Score & Term & Score & Term & Score \\
\hline
express & 0.7686 & thread & 1.0811 & compression & 1.0963 & delivery & 1.0899 \\
\hline
net & 0.763 & kingdom & 0.9953 & mouse & 1.0841 & circuit & 1.0893 \\
\hline
grid & 0.7314 & united & 0.8992 & assembly & 0.9898 & stable & 1.049 \\
\hline
deal & 0.7037 & net & 0.8177 & native & 0.9866 & sun & 1.0488 \\
family & 0.6934 & background & 0.7982 & controlled & 0.9639 & single & 1.0144 \\
\hline
united & 0.6925 & uniform & 0.7864 & root & 0.9551 & state & 1.0122 \\
\hline
primary & 0.6902 & translation & 0.771 & internal & 0.911 & deep & 0.9946 \\
\hline
reverse & 0.6855 & natural & 0.7427 & spectrum & 0.8867 & drive & 0.9844 \\
\hline
universal & 0.6677 & record & 0.7376 & bank & 0.8825 & free & 0.9687 \\
\hline
life & 0.6665 & stand & 0.7167 & derivative & 0.8823 & active & 0.9612 \\
\hline
\multicolumn{2}{|c|}{\vdots} & \multicolumn{2}{c|}{\vdots} & \multicolumn{2}{c|}{\vdots} & \multicolumn{2}{c|}{\vdots} \\
\hline
typical & 0.1355 & notion & 0.1569 & michael & 0.1601 & occurs & 0.2409 \\
\hline
specialized & 0.1286 & leave & 0.1523 & government & 0.1601 & interview & 0.2356 \\
\hline
richard & 0.1272 & appeared & 0.149 & david & 0.1568 & daniel & 0.2353 \\
\hline
compatible & 0.1257 & david & 0.1464 & political & 0.1559 & measured & 0.2277 \\
\hline
detect & 0.125 & nature & 0.1415 & required & 0.1539 & located & 0.2246 \\
\hline
benefit & 0.1223 & requiring & 0.1395 & dependent & 0.1497 & prevent & 0.2225 \\
\hline
infrastructure & 0.115 & corresponds & 0.1382 & charles & 0.1453 & increase & 0.1992 \\
\hline
corresponding & 0.1145 & authority & 0.1343 & james & 0.1417 & federal & 0.1885 \\
\hline
david & 0.112 & required & 0.1321 & peter & 0.1361 & growth & 0.1876 \\
\hline
authority & 0.109 & corresponding & 0.1176 & robert & 0.1271 & robert & 0.1866 \\
\hline
\end{tabular}

\medskip

\begin{tabular}{|lc|lc|lc|}
\hline
\multicolumn{2}{|c|}{Medical Device} & \multicolumn{2}{c|}{Medical Robot} & \multicolumn{2}{c|}{Sports Rehab Machine} \\
\hline
Term & Score & Term & Score & Term & Score \\
\hline
root & 0.8802 & stroke & 0.919 & kingdom & 0.907 \\
\hline
mouse & 0.8633 & kingdom & 0.893 & stroke & 0.8495 \\
\hline
kingdom & 0.8383 & vessel & 0.8651 & progressive & 0.8414 \\
\hline
iron & 0.8381 & thread & 0.8385 & net & 0.8334 \\
\hline
internal & 0.8043 & floating & 0.8045 & suspension & 0.8322 \\
\hline
progressive & 0.7957 & strain & 0.8018 & induction & 0.8244 \\
\hline
agent & 0.7875 & mouse & 0.7997 & thread & 0.8236 \\
\hline
express & 0.7733 & progressive & 0.7983 & root & 0.8093 \\
\hline
plasma & 0.7685 & die & 0.787 & transmission & 0.7871 \\
\hline
net & 0.7678 & secondary & 0.786 & die & 0.7821 \\
\hline
\multicolumn{2}{|c|}{\vdots} & \multicolumn{2}{c|}{\vdots} & \multicolumn{2}{c|}{\vdots} \\
\hline
argued & 0.1631 & corresponding & 0.1695 & corresponding & 0.196 \\
\hline
richard & 0.1608 & corresponds & 0.1693 & told & 0.1958 \\
\hline
authority & 0.1606 & feel & 0.167 & joseph & 0.1956 \\
\hline
michael & 0.1569 & coating & 0.1666 & understanding & 0.1953 \\
\hline
required & 0.154 & wife & 0.1603 & love & 0.1902 \\
\hline
peter & 0.1388 & michael & 0.159 & economic & 0.1902 \\
\hline
robert & 0.1381 & authority & 0.156 & coating & 0.1892 \\
\hline
david & 0.1201 & peter & 0.1437 & pay & 0.1856 \\
\hline
james & 0.1188 & david & 0.1412 & authority & 0.1838 \\
\hline
charles & 0.1157 & required & 0.1382 & causing & 0.1822 \\
\hline
\end{tabular}
\end{table*}

In order to study the cases of disagreement between the approaches proposed by this paper and Ferrari \textit{et al.} (2018), the top-5 words with the largest absolute differences between the assigned ranks have been reported for each scenario by Table \ref{table:diff}. The number of target words for each project scenario have also been mentioned in parenthesis. It can be observed that most of the cases of disagreement have a higher rank, i.e. relatively lower ambiguity score assigned by the linear transformation approach proposed by this paper. Most of such cases are proper names such as \textit{robert}, \textit{peter}, and \textit{daniel}. This is because of the problems associated with local-context approaches discussed in Section \ref{sec:motivation}, and the low ambiguity score given to such words by the proposed approach is in line with the expected behavior of a global-context approach.

\begin{table*}
\caption{Cases of Disagreement between the Linear Transformation approach ($R_1$) and the one suggested by Ferrari \textit{et al.} (2018) ($R_2$)}\label{table:diff}
\centering
\begin{tabular}{|lccc|lccc|lccc|}
\hline
\multicolumn{4}{|c|}{Light Controller (986)} & \multicolumn{4}{c|}{Mechanical CAD (1016)} & \multicolumn{4}{c|}{Medical Software (742)} \\
\hline
Term & $R_1$ & $R_2$ & $|R_1 - R_2|$ & Term & $R_1$ & $R_2$ & $|R_1 - R_2|$ & Term & $R_1$ & $R_2$ & $|R_1 - R_2|$\\
\hline
robert & 972 & 31 & 941 & notion & 1007 & 18 & 989 & peter & 741 & 12 & 729 \\
\hline
michael & 933 & 4 & 929 & third & 31 & 1008 & 977 & thomas & 727 & 14 & 713 \\
\hline
phenomenon & 971 & 60 & 911 & kingdom & 2 & 975 & 973 & third & 18 & 730 & 712 \\
\hline
peter & 902 & 8 & 894 & richard & 964 & 11 & 953 & richard & 712 & 19 & 693 \\
\hline
wide & 45 & 939 & 894 & green & 40 & 987 & 947 & mind & 707 & 38 & 669 \\
\hline
\end{tabular}

\medskip

\begin{tabular}{|lccc|lccc|lccc|lccc|}
\hline
\multicolumn{4}{|c|}{Athletes Network (569)} & \multicolumn{4}{|c|}{Medical Device (1168)} & \multicolumn{4}{c|}{Medical Robot (1507)} & \multicolumn{4}{c|}{Sports Rehab Machine (1624)} \\
\hline
Term & $R_1$ & $R_2$ & $|R_1 - R_2|$ & Term & $R_1$ & $R_2$ & $|R_1 - R_2|$ & Term & $R_1$ & $R_2$ & $|R_1 - R_2|$ & Term & $R_1$ & $R_2$ & $|R_1 - R_2|$ \\
\hline
daniel & 562 & 1 & 561 & peter & 1164 & 23 & 1141 & paul & 1486 & 8 & 1478 & daniel & 1613 & 3 & 1610 \\
\hline
robert & 569 & 17 & 552 & third & 56 & 1162 & 1106 & coating & 1501 & 29 & 1472 & love & 1619 & 17 & 1602 \\
\hline
main & 28 & 537 & 509 & richard & 1160 & 76 & 1084 & peter & 1505 & 38 & 1467 & peter & 1591 & 16 & 1575 \\
\hline
effect & 522 & 15 & 507 & white & 74 & 1150 & 1076 & third & 42 & 1504 & 1462 & coating & 1621 & 66 & 1555 \\
\hline
child & 59 & 558 & 499 & chess & 1102 & 39 & 1063 & richard & 1482 & 22 & 1460 & told & 1616 & 73 & 1543 \\
\hline
\end{tabular}
\end{table*}

\section{Conclusion and Future Work}
\label{sec:final}
Ambiguous requirements are a major hindrance to successful software development and it is necessary to avoid them from the elicitation phase itself. Although this problem has been studied extensively, cross-domain ambiguity has attracted research only in recent times. This paper proposes a global-context approach which makes use of linear transformation to map various domain-specific language models into a unified embedding space, allowing comparison of word embeddings trained from different corpora. It provides a logically effective way of determining potentially ambiguous words and a qualitative comparison with an existing local-context approach produces promising results. The planned future work includes a systematic quantitative evaluation of the proposed approach, extending the approach to consider multi-word phrases, defining an ambiguity threshold, and identifying better corpora sources.

\bibliographystyle{IEEEtran}\bibliography{IEEEabrv,refs}

\begin{thebibliography}{10}
\providecommand{\url}[1]{#1}
\csname url@samestyle\endcsname
\providecommand{\newblock}{\relax}
\providecommand{\bibinfo}[2]{#2}
\providecommand{\BIBentrySTDinterwordspacing}{\spaceskip=0pt\relax}
\providecommand{\BIBentryALTinterwordstretchfactor}{4}
\providecommand{\BIBentryALTinterwordspacing}{\spaceskip=\fontdimen2\font plus
\BIBentryALTinterwordstretchfactor\fontdimen3\font minus
  \fontdimen4\font\relax}
\providecommand{\BIBforeignlanguage}[2]{{%
\expandafter\ifx\csname l@#1\endcsname\relax
\typeout{** WARNING: IEEEtran.bst: No hyphenation pattern has been}%
\typeout{** loaded for the language `#1'. Using the pattern for}%
\typeout{** the default language instead.}%
\else
\language=\csname l@#1\endcsname
\fi
#2}}
\providecommand{\BIBdecl}{\relax}
\BIBdecl

\bibitem{pressman2010software}
R.~S. Pressman, \emph{Software Engineering: A Practitioner's Approach}.\hskip
  1em plus 0.5em minus 0.4em\relax McGraw-Hill, 2010.

\bibitem{aggarwal2005software}
K.~Aggarwal and Y.~Singh, \emph{Software Engineering}.\hskip 1em plus 0.5em
  minus 0.4em\relax New Age International (P) Limited, 2005.

\bibitem{singh2012object}
Y.~Singh and R.~Malhotra, \emph{Object-Oriented Software Engineering}.\hskip
  1em plus 0.5em minus 0.4em\relax PHI Learning, 2012.

\bibitem{wang2013}
Y.~Wang, I.~L. Manotas~Guti{\`e}rrez, K.~Winbladh, and H.~Fang, ``Automatic
  detection of ambiguous terminology for software requirements,'' in
  \emph{Natural Language Processing and Information Systems}.\hskip 1em plus
  0.5em minus 0.4em\relax Berlin, Heidelberg: Springer Berlin Heidelberg, 2013,
  pp. 25--37.

\bibitem{sch2019}
D.~Schlechtweg, A.~Hätty, M.~Del~Tredici, and S.~Schulte Im~Walde, ``A wind of
  change: Detecting and evaluating lexical semantic change across times and
  domains,'' in \emph{Proceedings of the 57th Annual Meeting of the Association
  for Computational Linguistics}, 01 2019, pp. 732--746.

\bibitem{ferrari2017}
A.~Ferrari, B.~Donati, and S.~Gnesi, ``Detecting domain-specific ambiguities:
  An {NLP} approach based on wikipedia crawling and word embeddings,'' in
  \emph{2017 IEEE 25th International Requirements Engineering Conference
  Workshops (REW)}, Sep. 2017, pp. 393--399.

\bibitem{mishra2019}
S.~{Mishra} and A.~{Sharma}, ``On the use of word embeddings for identifying
  domain specific ambiguities in requirements,'' in \emph{2019 IEEE 27th
  International Requirements Engineering Conference Workshops (REW)}, Sep.
  2019, pp. 234--240.

\bibitem{ferrari2018}
A.~Ferrari, A.~Esuli, and S.~Gnesi, ``Identification of cross-domain ambiguity
  with language models,'' in \emph{2018 5th International Workshop on
  Artificial Intelligence for Requirements Engineering (AIRE)}, Aug 2018, pp.
  31--38.

\bibitem{ferrari2019}
A.~Ferrari and A.~Esuli, ``An {NLP} approach for cross-domain ambiguity
  detection in requirements engineering,'' \emph{Automated Software
  Engineering}, vol.~26, no.~3, pp. 559--598, Sep 2019.

\bibitem{toews}
\BIBentryALTinterwordspacing
D.~Toews and L.~V. Holland, ``Determining domain-specific differences of
  polysemous words using context information,'' in \emph{Joint Proceedings of
  {REFSQ-2019} Workshops, Doctoral Symposium, Live Studies Track, and Poster
  Track co-located with the 25th International Conference on Requirements
  Engineering: Foundation for Software Quality {(REFSQ} 2019), Essen, Germany,
  March 18th, 2019.}, 2019. [Online]. Available:
  \url{http://ceur-ws.org/Vol-2376/NLP4RE19\_paper02.pdf}
\BIBentrySTDinterwordspacing

\bibitem{li2015}
Y.~Li, L.~Xu, F.~Tian, L.~Jiang, X.~Zhong, and E.~Chen, ``Word embedding
  revisited: A new representation learning and explicit matrix factorization
  perspective,'' in \emph{Proceedings of the 24th International Conference on
  Artificial Intelligence}, ser. IJCAI'15.\hskip 1em plus 0.5em minus
  0.4em\relax AAAI Press, 2015, pp. 3650--3656.

\bibitem{harris1954}
Z.~S. Harris, ``Distributional structure,'' \emph{Word}, vol.~10, no. 2-3, pp.
  146--162, 1954.

\bibitem{mikolov2013}
T.~Mikolov, I.~Sutskever, K.~Chen, G.~Corrado, and J.~Dean, ``Distributed
  representations of words and phrases and their compositionality,'' in
  \emph{Proceedings of the 26th International Conference on Neural Information
  Processing Systems - Volume 2}, ser. NIPS'13.\hskip 1em plus 0.5em minus
  0.4em\relax USA: Curran Associates Inc., 2013, pp. 3111--3119.

\bibitem{mikolov2013exploiting}
\BIBentryALTinterwordspacing
T.~Mikolov, Q.~V. Le, and I.~Sutskever, ``Exploiting similarities among
  languages for machine translation,'' \emph{ArXiv}, vol. abs/1309.4168, 2013.
  [Online]. Available: \url{http://arxiv.org/abs/1309.4168}
\BIBentrySTDinterwordspacing

\bibitem{hamilton}
W.~L. Hamilton, J.~Leskovec, and D.~Jurafsky, ``Diachronic word embeddings
  reveal statistical laws of semantic change,'' in \emph{Proceedings of the
  54th Annual Meeting of the Association for Computational Linguistics (Volume
  1: Long Papers)}.\hskip 1em plus 0.5em minus 0.4em\relax Berlin, Germany:
  Association for Computational Linguistics, Aug. 2016, pp. 1489--1501.

\bibitem{kulkarni}
V.~Kulkarni, R.~Al-Rfou, B.~Perozzi, and S.~Skiena, ``Statistically significant
  detection of linguistic change,'' in \emph{Proceedings of the 24th
  International Conference on World Wide Web}, ser. WWW '15.\hskip 1em plus
  0.5em minus 0.4em\relax Republic and Canton of Geneva, Switzerland:
  International World Wide Web Conferences Steering Committee, 2015, pp.
  625--635.

\bibitem{muromagi}
A.~Murom{\"a}gi, K.~Sirts, and S.~Laur, ``Linear ensembles of word embedding
  models,'' in \emph{Proceedings of the 21st Nordic Conference on Computational
  Linguistics}.\hskip 1em plus 0.5em minus 0.4em\relax Gothenburg, Sweden:
  Association for Computational Linguistics, May 2017, pp. 96--104.

\bibitem{artetxe2016}
M.~Artetxe, G.~Labaka, and E.~Agirre, ``Learning principled bilingual mappings
  of word embeddings while preserving monolingual invariance,'' in
  \emph{Proceedings of the 2016 Conference on Empirical Methods in Natural
  Language Processing}.\hskip 1em plus 0.5em minus 0.4em\relax Austin, Texas:
  Association for Computational Linguistics, Nov. 2016, pp. 2289--2294.

\bibitem{xing2015}
C.~Xing, D.~Wang, C.~Liu, and Y.~Lin, ``Normalized word embedding and
  orthogonal transform for bilingual word translation,'' in \emph{Proceedings
  of the 2015 Conference of the North {A}merican Chapter of the Association for
  Computational Linguistics: Human Language Technologies}.\hskip 1em plus 0.5em
  minus 0.4em\relax Denver, Colorado: Association for Computational
  Linguistics, May{--}Jun. 2015, pp. 1006--1011.

\bibitem{Sch1966}
P.~H. Sch{\"o}nemann, ``A generalized solution of the orthogonal procrustes
  problem,'' \emph{Psychometrika}, vol.~31, no.~1, pp. 1--10, Mar 1966.

\bibitem{tredici-fernandez-2017-semantic}
\BIBentryALTinterwordspacing
M.~D. Tredici and R.~Fern{\'a}ndez, ``Semantic variation in online communities
  of practice,'' in \emph{{IWCS} 2017 - 12th International Conference on
  Computational Semantics - Long papers}, 2017. [Online]. Available:
  \url{https://www.aclweb.org/anthology/W17-6804}
\BIBentrySTDinterwordspacing

\bibitem{hatty-etal-2019-surel}
A.~H{\"a}tty, D.~Schlechtweg, and S.~Schulte~im Walde, ``{SUR}el: A gold
  standard for incorporating meaning shifts into term extraction,'' in
  \emph{Proceedings of the Eighth Joint Conference on Lexical and Computational
  Semantics (*{SEM} 2019)}.\hskip 1em plus 0.5em minus 0.4em\relax Minneapolis,
  Minnesota: Association for Computational Linguistics, Jun. 2019, pp. 1--8.

\end{thebibliography}

\end{document}